%% file: main.tex
\documentclass[runningheads]{llncs}

\usepackage{algorithm,algpseudocode}
\usepackage{amsmath}
\usepackage{hyperref}
\usepackage{xcolor}
\usepackage[T1]{fontenc}

\usepackage{graphicx,verbatim}
\usepackage{subcaption}
\usepackage{booktabs}
\usepackage{multirow}
\usepackage{amssymb}

\input{macros.tex}

\begin{document}

\title{SCORPION: Addressing Scanner-Induced Variability in Histopathology}

\author{
Jeongun Ryu\inst{1} \and
Heon Song\inst{1} \and
Seungeun Lee\inst{1} \and
Soo Ick Cho\inst{1} \and
Jiwon Shin\inst{1} \and
Kyunghyun Paeng\inst{1} \and
Sérgio Pereira\inst{1}
}
\institute{Lunit Inc.\\
\email{\{rjw0205, heon.song, lsee1113, sooickcho, jwshin, khpaeng, sergio\}@lunit.io}
}

\maketitle

\input{sections/0_abstract}
\input{sections/1_introduction}
\input{sections/2_dataset}
\input{sections/3_method}
\input{sections/4_experiment}
\input{sections/5_conclusion}

\bibliographystyle{splncs04}
\bibliography{bibliography}

\end{document}

%% file: macros.tex
\newcommand{\score}[2]{{#1}\scriptsize{$\pm$#2}}

\makeatletter
\renewcommand\paragraph{\@startsection{paragraph}{4}{\z@}%
                                    {1.00ex \@plus1ex \@minus.2ex}%
                                    {-1em}%
                                    {\normalfont\normalsize\bfseries}}

\makeatother

%% file: sections/0_abstract.tex
\begin{abstract}
Ensuring reliable model performance across diverse domains is a critical challenge in computational pathology. 
A particular source of variability in Whole-Slide Images is introduced by differences in digital scanners, thus calling for better scanner generalization.
This is critical for the real-world adoption of computational pathology, where the scanning devices may differ per institution or hospital, and the model should not be dependent on scanner-induced details, which can ultimately affect the patient's diagnosis and treatment planning. 
However, past efforts have primarily focused on standard domain generalization settings, evaluating on unseen scanners during training, without directly evaluating consistency across scanners for the same tissue.
To overcome this limitation, we introduce SCORPION, a new dataset explicitly designed to evaluate model reliability under scanner variability. 
SCORPION includes 480 tissue samples, each scanned with 5 scanners, yielding 2,400 spatially aligned patches. 
This scanner-paired design allows for the isolation of scanner-induced variability, enabling a rigorous evaluation of model consistency while controlling for differences in tissue composition.
Furthermore, we propose SimCons, a flexible framework that combines augmentation-based domain generalization techniques with a consistency loss to explicitly address scanner generalization. 
We empirically show that SimCons improves model consistency on varying scanners without compromising task-specific performance.
By releasing the SCORPION dataset\footnote{SCORPION dataset is available at \href{https://doi.org/10.5281/zenodo.16517924}{https://doi.org/10.5281/zenodo.16517924}} and proposing SimCons, we provide the research community with a crucial resource for evaluating and improving model consistency across diverse scanners, setting a new standard for reliability testing. 

\keywords{Computational Pathology \and Scanner Generalization.}
\end{abstract}

%% file: sections/1_introduction.tex
\section{Introduction}
Computational pathology \cite{cpath} has been increasingly adopted in modern healthcare, offering high-throughput and precise analysis of histopathological samples for disease diagnosis, prognosis~\cite{park2023tumor}, and treatment planning~\cite{lee2022artificial}. 
Advances in computational models have revolutionized traditional workflows, significantly enhancing diagnostic accuracy and efficiency~\cite{dbb27c369cda4e349799fba80ac467b0,ocelot}. 
However, a critical challenge hindering the deployment of these models in real-world clinical settings is the variability introduced by different scanners used to digitize tissue slides into whole-slide images (WSI).

\begin{figure}[t]
    \centering
    \hfill
        \includegraphics[width=\linewidth]{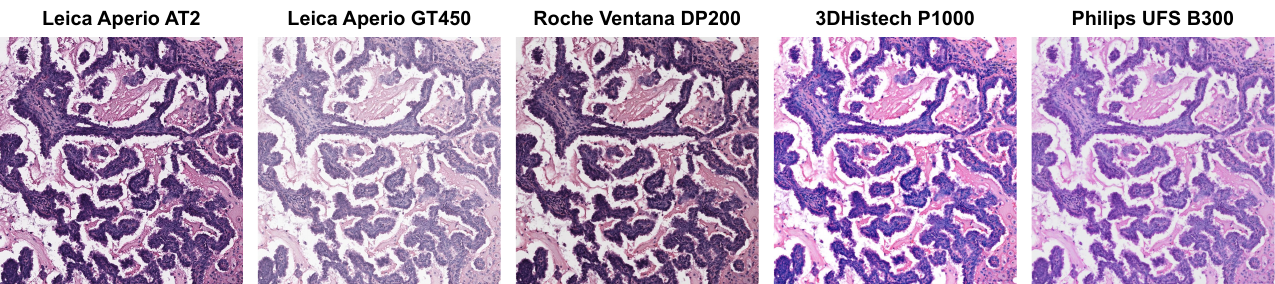}
    \hfill
    \vskip -2mm
    \caption{\textbf{Example of scanner-induced variability in SCORPION.} The same tissue region is digitized using five different scanners. Despite capturing identical histological structures, variations in color, contrast, and texture highlight the challenges of inter-scanner variability in histopathology. 
    }
    \label{fig:data_sample}
    \vspace{-3mm}
\end{figure}

Scanners vary significantly in the manufacturer, hardware, and image acquisition settings, introducing scanner-induced variability that can affect WSI, even for the same tissue slide (see Fig. \ref{fig:data_sample}). While human pathologists can intuitively disregard such differences, computational models are susceptible to these discrepancies, leading to inconsistent predictions \cite{midog2021,cameleyon17}. 
This is particularly problematic in clinical workflows, where inconsistent predictions could result in varying diagnoses or treatment recommendations, jeopardizing patient outcomes \cite{duenweg2023whole}. 

Prior efforts, including initiatives such as Camelyon17 \cite{cameleyon17} and MIDOG2021 \cite{midog2021}, have contributed benchmarks for domain generalization in histopathology. However, these datasets do not provide scanner-paired images, thus making it impossible to directly assess inter-scanner consistency. 
More recently, datasets \cite{ochi2024plism,wilm2023multiscanner,cosas2024} have introduced paired scans of the same tissue acquired with different scanners. While these datasets enable the possibility of consistency evaluation, prior work has predominantly adopted conventional domain generalization settings, evaluating on unseen scanners during training, without explicitly analyzing consistency across scanners for the same tissue. In contrast, we directly leverage the scanner-paired nature of the data to rigorously evaluate inter-scanner consistency, which is critical to ensure clinical reliability of the model.

Building upon these recent datasets that include scanner-paired images, we introduce SCORPION, a novel H\&E dataset designed explicitly to evaluate model consistency under scanner variability. SCORPION comprises 480 tissue regions, each scanned using 5 scanners, resulting in 2,400 spatially aligned patches. By isolating scanner variability from tissue heterogeneity, SCORPION enables rigorous evaluation of model consistency across scanners. This dataset establishes a new benchmark for scanner generalization, facilitating the development of robust and reliable computational models. Therefore, SCORPION will help accelerate research in this critical area.

Further, we propose SimCons, a flexible framework that explicitly incorporates consistency as a core objective. 
SimCons combines style-based augmentation with a consistency loss to encourage consistent predictions on style variation. 
By emphasizing consistency, we address the unique challenges posed by scanner variability, ensuring the clinical reliability of the model without compromising task-specific performance. 
Together, SCORPION and SimCons provide a critical resource and methodology for advancing scanner generalization research, setting a new standard for reliability testing in CPath.

In summary, our contributions are threefold: (1) we publicly release SCORPION, a comprehensive dataset of spatially aligned patches scanned using 5 different scanners; (2) we establish a novel problem setting and evaluation protocol specifically addressing inter-scanner variability; and (3) we introduce SimCons, a straightforward yet effective framework that mitigates inter-scanner variability while maintaining or enhancing task-specific performance.

%% file: sections/2_dataset.tex
\section{SCORPION dataset}

\label{sec:dataset}

SCORPION is an H\&E-stained histopathology dataset specifically built to enable the evaluation of model consistency across scanners. 
Each sample of the SCORPION is composed of 5 patches, $(x_{AT2}, x_{GT450}, x_{DP200}, x_{P1000}, x_{B300})$, where patches in a sample share the same tissue content, differing only in the scanner used.
Fig.\ref{fig:data_sample} shows a sample of the SCORPION dataset.

\subsection{Dataset Collection.} 
We collected 48 H\&E-stained tissue slides, where each slide was digitized using the following scanners: \textit{Leica Aperio AT2, Leica Aperio GT450, Roche Ventana DP200, 3DHistech P1000,} and \textit{Philips UFS B300}. 
To ensure spatial consistency across the scans, we aligned the images of the same tissue slide through a registration algorithm based on ORB \cite{orb} feature extraction and affine transformation. 
From each aligned slide, we extracted 10 regions, each measuring 800$\mu$$m$ $\times$ 800$\mu$$m$, and resized them as 1024 $\times$ 1024 pixel patches. 
This process resulted in 480 samples, each consisting of 5 patches from the same tissue region but captured by different scanners, leading to a total of 2,400 patches.

\subsection{Dataset Analysis.} 
\label{sec:dataset_analysis}
\begin{figure}[t]
    \centering
    \begin{subfigure}[b]{\textwidth}
        \centering
        \includegraphics[width=1\textwidth]{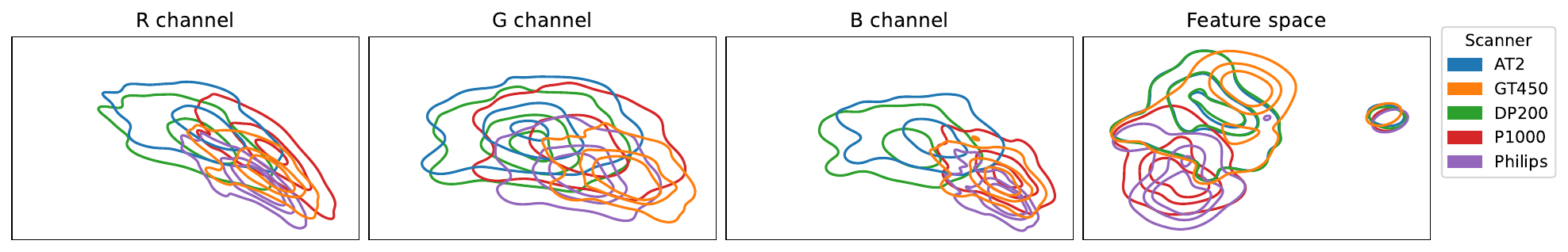}
        \caption{Unpaired Analysis.}
        \label{fig:unpaired}
    \end{subfigure}
    \begin{subfigure}[b]{\textwidth}
        \centering
        \includegraphics[width=1\textwidth]{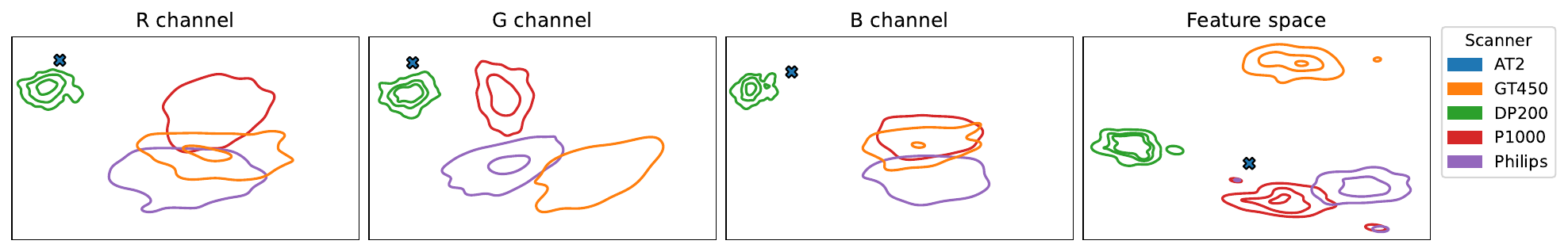}
        \caption{Paired Analysis.}
        \label{fig:paired}
    \end{subfigure}

    \caption{\textbf{Input- and feature-level analysis across scanners in SCORPION.} Density contours highlight the distribution of patches across different scanners. In the left three columns, (x,y) denote the mean and standard deviation of pixel values for each RGB channel. In the last column, features extracted using a ResNet50 pre-trained on ImageNet are projected into a 2D space using UMAP. (a) In unpaired analysis, all patches are plotted without considering the paired-patch relationship. (b) On the other hand, in paired analysis, AT2 (blue) is set as the reference scanner, and deviations are computed for each paired patch relative to its corresponding AT2 patch, positioning all AT2 patches at (0,0).}
    \vspace{-2mm}
    \label{fig:dataset_analysis}
\end{figure}

To analyze scanner-induced variability in the SCORPION dataset, we first investigate the input-level distributions.
For each RGB channel, the mean and standard deviation of the pixel values are computed for each patch, which quantifies the distribution of pixel intensities across scanners. 
Then, we further investigate scanner-induced differences in the feature space. Features are extracted for each patch using ResNet50 \cite{He2015}, pre-trained on ImageNet \cite{imagenet}. 
For visualization, we project the features into a 2D space by UMAP \cite{umap}. The density contours of each scanner are displayed in Fig.~\ref{fig:dataset_analysis}.

\paragraph{Unpaired analysis} We begin with an \textit{unpaired analysis}, where the paired-patch relationship is ignored, and all patches are plotted with scanner labels assigned. As shown in Fig.~\ref{fig:unpaired}, it reveals substantial overlap between scanners, making it challenging to identify scanner-specific characteristics, potentially leading to mistakenly consider there are no major inter-scanner differences.

\paragraph{Paired analysis}
To overcome the limitations of the \textit{unpaired analysis}, we leverage the paired nature of the SCORPION dataset for a \textit{paired analysis}.
In this analysis, AT2 is selected as the reference scanner, and deviations are computed for each paired patch relative to the corresponding AT2 patch, ensuring that all AT2 patches are positioned at (0,0) in Fig.~\ref{fig:paired}.
As a result, we found that the distributions of each scanner are easily separated in feature space, which indicates that the pre-trained encoder captures scanner-specific information from the input.
Also in the RGB channels, except for the GT450 (orange), all the scanner distributions are well separated.
This shows the superiority of \textit{paired analysis}, which provides a clearer view of scanner-induced variability compared to \textit{unpaired analysis}, by utilizing the scanner-paired structure.

\subsection{Inter-Scanner Consistency Evaluation Protocol.}
\label{sec:evaluation_protocol}

Prior efforts \cite{cosas2024,midog2021,cameleyon17,ochi2024plism,wilm2023multiscanner} that studied inter-scanner variability have typically followed the standard domain generalization protocol by evaluating the model on unseen scanners during training. However, this evaluation approach does not assess whether the model produces consistent outputs when only the scanner changes while the underlying tissue remains the same. In real-world clinical settings, where scanner types can vary between hospitals, this limitation poses a significant risk: a patient could receive different clinical decisions depending solely on the scanner used.  Motivated by this, we leverage the scanner-paired design of SCORPION to propose a new evaluation protocol that rigorously quantifies model consistency across scanners, addressing a critical gap in existing evaluation methods.

In this evaluation protocol, we compute the consistency score (e.g. Dice score in tissue segmentation task) between predictions from scanner-paired patches, for each scanner pair.
Since SCORPION includes 5 scanners, there are 10 unique scanner pairs, resulting in 10 consistency scores.
Finally, we compute two measurements: the average and minimum of the 10 consistency scores. 
The average score represents the overall consistency of the model across different scanner pairs, while the minimum score captures the lower bound, ensuring that the model maintains a certain level of consistency, regardless of which scanner pair is used. 
This minimum metric is particularly important for assessing the worst-case scenario in real-world clinical settings, where scanner-induced variability can affect medical treatment.

%% file: sections/3_method.tex
\section{Method}

\begin{figure}[t]
    \centering    
    \includegraphics[width=\textwidth]{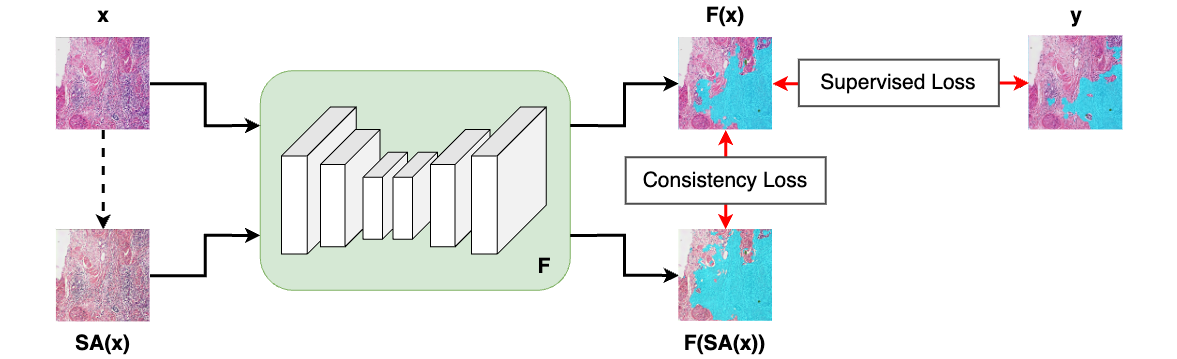}
    \caption{\textbf{Overview of SimCons framework.} SimCons improves model consistency across scanners by integrating style-based augmentation (SA) with a consistency loss. The SA module transforms a training image into a style-altered version while preserving its content. Both the original and style-altered images are fed into the model ($F$), and their predictions are aligned using a consistency loss. Simultaneously, a supervised loss ensures task-specific learning.}
    \label{fig:method}
\end{figure}

To enhance the model consistency across diverse scanners, we propose SimCons, a simple framework combining style-based augmentation (SA) with a consistency loss. 
SimCons leverages an SA to synthesize style-altered images while preserving the content of the original image. 
Then, the consistency loss explicitly encourages consistent model predictions between original and style-altered images. An overview of the SimCons framework is illustrated in Fig.\ref{fig:method}.

\subsection{Style-based Augmentation (SA)}
Due to the differences among scanners, including optical components and sensor characteristics, each scanner introduces unique color profiles, texture patterns, and contrast properties. As a result, the same tissue scanned by multiple scanners can have varying styles for the same content.
In prior work, various SA methods \cite{pmlr-v143-faryna21a,Shen_2022,Tellez2018HAE,yang2021skunetmodelfourierdomain} were studied to improve the robustness to these style variations. 
In Tab.\ref{tab:main}, we observe that SA methods improve the model consistency on scanner variations while maintaining or improving performance in primary tasks.

\subsection{Consistency Loss}
While SA can enhance model performance on unseen scanners by generating images with various styles during training, it alone cannot ensure consistent predictions across scans with identical tissue content. 
To explicitly encourage the model to produce consistent output when only the scanner changes, we apply a consistency loss between the model’s predictions on the original image and its style-altered counterpart. 
The final loss of SimCons aggregates the supervised and consistency losses, formulated as follows:

\begin{equation}
    \mathcal{L}_{\text{total}} = \mathcal{L}_{\text{supervised}}(F(x), y) + \lambda \cdot \mathcal{L}_{\text{consistency}}(F(x), F(\text{SA}(x)))
    \label{equation:final_loss}
\end{equation}

\noindent where $(x,y)$ represents a training patch with a corresponding label, and $F$ denotes the model to be trained.
The hyperparameter $\lambda$ controls the impact of the consistency loss.
Intuitively, a higher $\lambda$ can reduce the scanner-induced variability of the model predictions. But, if set too large, it can hinder the primary task performance.
We conduct an in-depth study on the effect of $\lambda$ in Section \ref{experiment:lambda_tradeoff}.

%% file: sections/4_experiment.tex
\section{Experiments}
\label{section:experiment}

In this section, we assess the model consistency over scanners and show the effectiveness of the proposed SimCons framework. 
We use tissue segmentation as the primary task; accordingly, Dice score is chosen for both consistency and primary task metrics.
For this, we first acquire tissue segmentation models by training and validating on the internal HTS dataset. 
Then, we measure two types of metrics: (1) scanner-paired Dice as a consistency metric following Section \ref{sec:evaluation_protocol} with the SCORPION dataset and (2) primary task Dice with the HTS dataset.

HTS is an internal dataset containing 8,327 patches extracted from 3,399 H\&E-stained WSIs, decomposed into training, validation, and test sets with 4,027, 3,947, and 353 patches, respectively.
Each patch is a 1024 $\times$ 1024 pixel image that represents a tissue region of approximately 800$\mu$$m$ $\times$ 800$\mu$$m$. Per-pixel manual annotations for semantic segmentation consider 3 classes: cancer area, cancer stroma, and background.
All the WSIs are scanned with \textit{Leica Aperio AT2} and \textit{3DHistech P1000}. Importantly, this is only a subset of the scanners used in the SCORPION dataset, thus allowing us to assess generalization. 

\subsection{Experimental Setup}
We use DeepLabV3+ \cite{deeplabv3plus2018} with a ResNet34 \cite{He2015} backbone for training the models on the HTS dataset. 
Models are trained for 150 epochs with the Adam optimizer \cite{KingmaB14}.
The Dice loss \cite{sudre2017generalised} is used for both supervised and consistency loss.
During training, four data augmentations are randomly applied, including two photometric (Gaussian blur, Gaussian noise) and two geometric (horizontal flip, vertical flip) transformations.
For each experiment, the best epoch is chosen based on the performance on the HTS validation set.
All experiments are repeated five times, and the mean and standard deviation of the metrics are reported. 

\input{tables/main_result}

\subsection{Main Results}
We adopt three SA methods with consistency loss to validate the efficacy of the SimCons framework. The adopted SA methods are the following:

\paragraph{ColorJitter.} It is a commonly used data augmentation technique that randomly alters the brightness, contrast, saturation, and hue of an image. 
Despite its simplicity, it has been proven to improve the generalizability of the model, especially in color-related variations \cite{TELLEZ2019101544}.

\paragraph{RandStainNA.} It is designed to address stain variations in histopathology images by combining stain normalization and stain augmentation \cite{Shen_2022}.

\paragraph{FDA.} It is an empirically proven technique to improve model generalization. It generates diverse style images by transferring low-frequency information between training images, utilizing the content-preserving characteristic of the Fourier phase component \cite{Yang_2020_CVPR}. \vspace{5pt}

From Tab.\ref{tab:main}, we find that all SA methods show improved consistency over scanners compared to the Baseline, which does not have a style-based augmentation.
Adding a consistency constraint on top of the SA methods further boosts the consistency metric. This implies that consistency loss is the key to enhancing the model consistency over scanners by enforcing the model to output style-invariant predictions.
In addition, adding a consistency constraint also improves the primary task performance, especially in the test set. 
This suggests that while the consistency loss is aimed at reducing scanner-induced variability, it also improves generalization, further showing the benefits of the SimCons framework.

\begin{figure}[t]
    \centering
    \includegraphics[width=\textwidth]{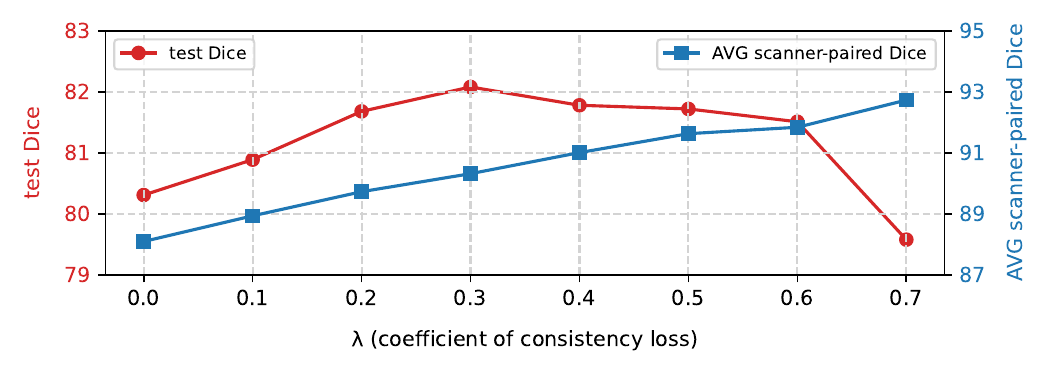}
    \caption{\textbf{Consistency vs. Primary Task.}
    Adjusting $\lambda$ can effectively balance improved scanner consistency with primary task performance.}
    \label{fig:ablation_loss_alpha}
\end{figure}

\subsection{Trade-off: Scanner Robustness vs Task Performance}
\label{experiment:lambda_tradeoff}
As described in Eq.\ref{equation:final_loss}, the loss function of SimCons consists of two terms: a supervised loss for learning the primary task and a consistency loss to encourage consistent predictions over style variation. 
The total loss is weighted with a coefficient $\lambda$ controlling the influence of the consistency loss.
In Fig.\ref{fig:ablation_loss_alpha}, we observe a trade-off between the primary task performance (red) and consistency score (blue).
When $\lambda = 0.0$, the model only considers the primary task, leading to a low consistency score. 
With increased $\lambda$, the consistency score improves consistently. 
Additionally, until $\lambda = 0.3$, we observe that it helps improving the primary task.
However, the performance of the primary task degrades for $\lambda > 0.3$. 
This is explained by a mode collapse phenomenon, where the model can reduce the consistency loss by generating a trivial but consistent prediction, thus harming the primary task's learning. 
Therefore, it is important to balance the supervised and consistency losses to achieve satisfactory consistency while showing strong performance in the primary task. 
This can be done by manipulating $\lambda$, where the optimal range was 0.3 to 0.5 within our experiment setting.

%% file: tables/main_result.tex
\begin{table}[t]
\centering

{
\setlength{\tabcolsep}{0.6em}

{
\caption {
\textbf{Tissue segmentation performance with consistency score.}
The Baseline has no style-based augmentation.}
\vspace{2mm}

\label{tab:main}
\begin{tabular}{ccccc}
    \toprule
    \multicolumn{1}{c}{} & \multicolumn{2}{c}{scanner-paired Dice} & \multicolumn{2}{c}{primary task Dice} \
    \\
    \cmidrule(lr){2-3}\cmidrule(lr){4-5}
    Method & \multicolumn{1}{c}{\textbf{Avg}} & \multicolumn{1}{c}{\textbf{Min}} & \multicolumn{1}{c}{\textbf{Val}} & \multicolumn{1}{c}{\textbf{Test}} 
    \\
    \midrule
    Baseline & \score{84.89}{0.56} & \score{75.63}{0.92} & \score{68.71}{0.06} & \score{80.04}{0.40} 
    \\
    \midrule
    ColorJitter & \score{86.77}{0.46} & \score{78.65}{0.85} & \score{68.93}{0.07} & \score{80.54}{0.42} 
    \\
    + SimCons & \score{89.64}{0.20} & \score{83.89}{0.40} & \score{69.45}{0.13} & \score{81.32}{0.26} 
    \\
    \midrule
    RandStainNA \cite{Shen_2022}  & \score{88.47}{0.37} & \score{81.93}{0.75} & \score{68.75}{0.06} & \score{80.56}{0.49} 
    \\
    + SimCons & \score{90.22}{0.21} & \score{84.92}{0.34} & \score{69.45}{0.14} & \score{81.75}{0.44} 
    \\
    \midrule
    FDA \cite{Yang_2020_CVPR} & \score{88.10}{0.32} & \score{81.98}{0.56} & \score{68.41}{0.05} & \score{80.31}{0.54} \\
    + SimCons & \score{90.32}{0.27} & \score{85.43}{0.35} & \score{68.87}{0.07} & \score{82.08}{0.28} \\
    \bottomrule
\end{tabular}
}
}
\end{table}

%% file: sections/5_conclusion.tex
\section{Conclusion}

This paper introduces SCORPION, a new dataset specifically designed to isolate scanner-induced variability in histopathology. 
By providing scanner-paired patches from five distinct scanners, SCORPION enables precise evaluation of model consistency across scanners while controlling for tissue differences. 
Furthermore, we propose an evaluation protocol that leverages SCORPION’s unique structure to quantify inter-scanner variability, establishing a new benchmark for scanner generalization research.
Beyond dataset contributions, we present SimCons, a flexible framework that integrates style-based augmentation with consistency loss to explicitly mitigate scanner-induced prediction variability. Our results demonstrate that SimCons enhances model consistency across scanners without compromising task performance.
These findings underscore the critical role of inter-scanner variability in real-world deployment, where reliable and consistent model predictions are essential for clinical decision-making. 
By releasing SCORPION and its evaluation framework, we aim to provide the research community with a foundational resource for improving model robustness and advancing the adoption of AI-driven pathology in clinical practice.

\subsubsection{Disclosure of Interests.} The
authors have no competing interests to declare that are relevant to the content of this article.